\documentclass[]{wam}
\usepackage[utf8]{inputenc}
\usepackage[T1]{fontenc}
\usepackage{url}
\usepackage{booktabs}
\usepackage{amsfonts}
\usepackage{amsmath}
\usepackage{amssymb}
\usepackage{microtype}
\usepackage{xcolor}
\usepackage{textcomp}
\usepackage{graphicx}
\usepackage{array}
\usepackage{multirow}
\usepackage{xspace}

\newcommand{\method}{WAM4D}
\providecommand{\keywords}[1]{\noindent\textbf{Keywords:} #1}

\title{WAM4D: Fast 4D World Action Model via \\ Spatial Register Tokens}

\author[1,2,*]{Ying Li}
\author[1,3,*]{Xiaobao Wei}
\author[1,3,\dagger]{Jiajun Cao}
\author[1]{Hao Wang}
\author[2]{Xiaowei Chi}
\author[1]{Chengyu Bai}
\author[1]{Qianpu Sun}
\author[1]{Jiajun Li}
\author[2]{Xiaojie Zhang}
\author[1]{Peidong Jia}
\author[3]{Jian Tang}
\author[2,\ddagger]{Sirui Han}
\author[1,3,\ddagger]{Shanghang Zhang}

\affiliation[1]{Peking University}
\affiliation[2]{The Hong Kong University of Science and Technology}
\affiliation[3]{Beijing Innovation Center of Humanoid Robotics}

\contribution[*]{Equal Contribution}
\contribution[\dagger]{Project Leader}
\contribution[\ddagger]{Equal Corresponding Author}

\abstract{
World action models (WAMs) have recently shown promise in jointly modeling future observations and executable robot actions. However, most existing WAMs still operate in 2D video or latent spaces, where visually plausible rollouts miss the 3D spatial constraints and occluded contact geometry required for precise manipulation. While geometric foundation models offer strong priors for recovering dense 3D structure and motion from visual observations, forcing WAMs to predict the dense 4D representation introduces costly geometric decoding and slows down causal action generation. To address the trade-off, we present \method{}, a fast 4D world action model that uses lightweight spatial register tokens as training-time future-depth readouts to transfer pretrained geometric priors into a causal video-action transformer, then removes the register branch for lightweight action inference. To prevent non-causal shortcuts, we further design causal mixture attention for the Mixture-of-Transformers (MoT) WAM backbone, defining modality-specific visibility among video, action, and geometry tokens. Comprehensive experiments on RoboTwin 2.0 and challenging real-world manipulation tasks show that \method{} improves spatial consistency and achieves competitive action prediction while maintaining efficient inference.
}

\date{\today}
\code{\url{https://github.com/myendless1/wam4d}}

\begin{document}
\thispagestyle{firstheader}
\maketitle

\keywords{World Action Model, 4D World Modeling, Robot Manipulation}

\section{Introduction}

World models have been viewed as a foundation for embodied intelligence by predicting how the physical environment evolves under interaction~\citep{wan2025wan, yang2025cogvideox, ding2025understanding}.
With the rapid progress of video generative models, this predictive capability has begun to transfer from general video modeling to embodied domains~\citep{chi2025wow, li2026manipdreamer, li2026manipdreamer3d, zeng2026rethinking, chen2026abot}.
As robotic simulators of future scene evolution, world models support data generation and policy evaluation~\citep{fan2026wow, zhou2024robodreamer, yu2025manigaussian, wu2026phymix}. The generated videos are further converted into executable actions using inverse dynamics models.

VLA policies have become a dominant route for robot control by directly mapping visual observations and language instructions to actions~\citep{rt1, pi0, pi05, cao2026fastdrivevla, pi07}.
However, manipulation is not a simple observation to action mapping.
It requires reasoning about scene geometry, contacts, and object dynamics.
Current VLAs learn these factors mainly from action supervision, which leads to weak generalization ability~\citep{zhang2026vlm4vla}.
Recent spatially grounded VLAs address this issue by injecting 3D inputs or geometric foundation priors into policy learning~\citep{qu2025spatialvla,li2026pointvla,sun2025geovla,li2025spatial,wang2026vega}.
These methods improve the spatial representation of direct policies, but future scene evolution and robot dynamics remain implicit in the action prediction objective.
Therefore, recent works have turned to world action models (WAMs).
WAMs extend world models from robotic video generation to joint prediction of future observations and executable robot actions~\citep{lingbotva, fastwam, team2026motubrain, zhang2026pelican}.
By using the priors of video generative foundation models, this end-to-end formulation couples imagination and control in one pipeline.
It learns an implicit transition model that predicts how the scene and robot may evolve from the initial observation and task instruction.

\begin{figure}[!t]
    \centering
    \includegraphics[page=1,width=1.0\linewidth]{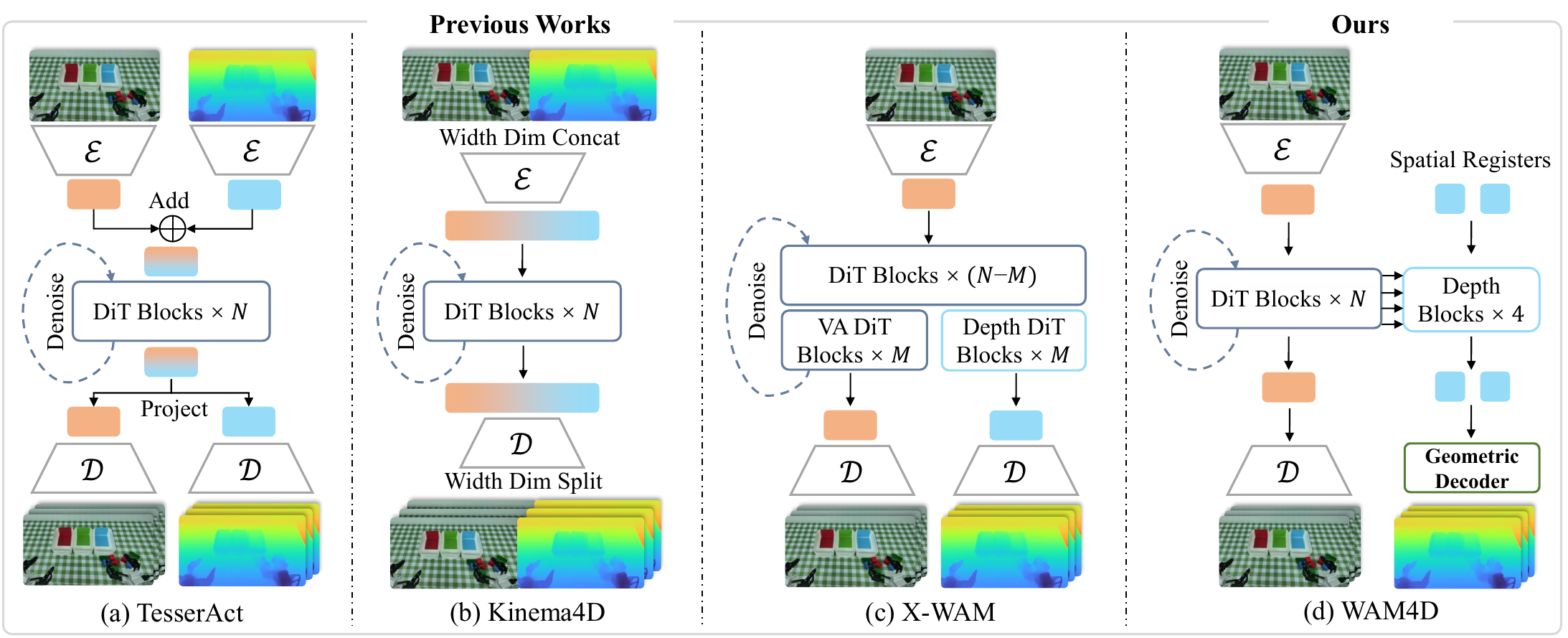}
    \caption{Comparison with representative 4D world-action modeling designs. TesserAct projects geometry into a 4D scene representation, Kinema4D concatenates geometric and visual tokens, and X-WAM uses modal adaptation for explicit RGB-D future synthesis. In contrast, \method{} distills geometric foundation priors through spatial register tokens, then removes the geometry readout for fast action inference. The inverse dynamics model in TesserAct and joint action denoising in X-WAM and \method{} are omitted for clarity.}
    \label{fig:teaser}
\end{figure}

Despite this progress, most WAMs represent future states in 2D video or latent spaces.
Such rollouts are useful for learning visual dynamics, but they provide an incomplete physical state for manipulation~\citep{ai2025review}.
Precise manipulation requires reasoning over object extent, occluded surfaces, free space, contacts, and robot motion across future steps.
A visually plausible rollout hides errors in contact geometry, which directly affect spatially precise robotic manipulation.
Recent 4D embodied world models address this limitation by augmenting future prediction with depth maps, surface normals, point clouds, or point maps~\citep{xwam, wang2026mvista, xu2026kinema4d, tesseract, tian2026starry}.
These representations make predicted futures more aligned with the physical state needed for action inference.
However, existing designs often treat dense 4D geometry as an explicit inference target.
These methods improve reconstruction fidelity but require dense geometry decoding or optimization during action inference, which increases deployment costs and latency.
More fundamentally, explicit 4D prediction may shift the WAM objective toward geometric reconstruction.
Current methods do not ensure that geometric priors strengthen the causal coupling between video prediction and action generation.

To address these issues, we propose \method{}, a fast 4D world action model that uses spatial register tokens as a compact bridge between 2D WAM latents and geometric priors. \method{} uses geometry as a training-time readout target rather than an additional sensory input or inference-time output: spatial registers query history video tokens, decode depth through a pretrained geometric head, and backpropagate the depth loss into the history video features used for action prediction. At deployment, the geometric head and depth readout are removed, so the policy keeps the same lightweight 2D observation-to-action interface.

Our main contributions are:
\begin{itemize}
    \item We propose \method{}, a fast 4D world action model that brings geometric foundation priors into a causal video-action model through a compact spatial-register interface.

    \item We design causal mixture attention for Mixture-of-Transformers (MoT). It defines visibility across video, action and spatial register tokens, enabling geometry supervision during training and lightweight inference at deployment.

    \item We evaluate \method{} on RoboTwin 2.0 and real-world long-horizon tasks. Experiments show improved spatial consistency and success rates while preserving efficient causal action inference.
\end{itemize}

\section{Related Work}

\noindent\textbf{Embodied World Models.}
Video generative models have made strong progress in modeling temporal visual dynamics~\citep{ho2022imagen, blattmann2023stable, wan2025wan, yang2025cogvideox, seedance2026seedance}.
This progress has motivated embodied world models that use video generation as learned simulators for interactive environments~\citep{chi2025wow, li2026manipdreamer, li2026manipdreamer3d, zeng2026rethinking, chen2026abot}.
Several works learn interactive world models from mixed embodied data and Internet videos~\citep{yang2023unisim, yin2026genie}.
Other works predict robotic videos and convert them into actions with inverse dynamics models~\citep{du2023learning, zhou2024robodreamer, wang2025language}.
These works show that video priors can model future scene evolution, but future prediction and action generation are separated.
Recent world action models (WAM) jointly model future observations and executable robot actions~\citep{wang2026world, motus, team2026motubrain, lingbotva, fastwam, xwam}.
% LingBot-VA introduces causal video-action modeling for jointly optimizing video prediction and action generation~\citep{}.
% Fast-WAM shows that video modeling during training can provide much of the WAM benefit without test-time future generation~\citep{fastwam}.
% X-WAM adds a dedicated depth branch by replicating the final DiT blocks for explicit RGB-D future synthesis~\citep{xwam}.
In contrast, \method{} uses lightweight spatial registers to transfer 4D geometric priors into video-action representations.

\noindent\textbf{Geometric Foundation Models.}
Explicit geometry provides important priors for embodied perception and control.
One line of work represents scenes with optimized 3D structures, such as neural radiance fields and 3D Gaussian primitives, which support view synthesis, planning, data generation, and manipulation~\citep{nerf,gaussian_splatting,gnfactor,manigaussian,splatmover,manipulateanything, huang2024s3g, wei2025emd, wei2026parkgaussian}.
Another line develops feed-forward geometric foundation models that recover depth, point maps, camera poses, and dense correspondence directly from images~\citep{wang2024dust3r, wang2025vggt, lin2025depth, wang2025pi, wei2025gazegaussian, wang2025embodiedocc++, wu2026feed, wang2026vggt}.
These models enable scalable geometric supervision from large-scale robot videos.
Recent embodied world models further inject geometry into future prediction by generating RGBD videos, normals, point maps, 3D trajectories, or 4D robot world interactions~\citep{tesseract, shen2026lyra, huang2026enerverse, yang2026neoverse, xu2026kinema4d, tian2026starry, tu2026embody4d}.
Collectively, these methods indicate that explicit geometry improves physical consistency and spatial reasoning.
However, they often require geometry to appear as an input, output, or reconstructed scene state during generation.
\method{} instead uses future depth as a training signal through spatial registers.

\noindent\textbf{Robotic Manipulation Policy.}
Robot manipulation policies commonly learn a direct mapping from observations to actions.
Early methods model action sequences from demonstrations~\citep{zhao2023learning, chi2025diffusion}.
Recent policies scale this paradigm with large language models.
RT series formulate robot control as sequence prediction over visual, language, and action tokens~\citep{rt1, rt2}.
Several works provide open generalist policies trained on large robot datasets~\citep{kim2024openvla, team2024octo, liu2026last}.
Pi series further improve cross embodiment generalization with larger training mixtures and stronger action modeling~\citep{liu2025rdt,pi0,pi05,pi07}.
Recent spatially grounded VLAs further improve action prediction by injecting geometric foundation priors~\citep{qu2025spatialvla,li2026pointvla,sun2025geovla,li2025spatial,wang2026vega}.
% These policies provide strong semantic grounding and scalable action prediction.
However, they usually learn geometry, contacts, and dynamics only through action labels. In contrast, WAMs turn video generative models into robot policies by jointly modeling future observations and executable actions.
% They provide a structured way to learn physical dynamics beyond direct action supervision.

% The deployed policy removes the geometric head and dense depth readout, while retaining a lightweight RGB action interface.

% \paragraph{Register tokens.}
% Register tokens were introduced in vision transformers as additional learnable slots that absorb global computation and improve dense visual feature maps~\citep{vitregisters}.
% \method{} repurposes this idea for embodied world modeling.
% Rather than adding registers to clean up a static visual encoder, we use spatially aligned registers as an auxiliary geometric readout interface.
% They query causal WAM states during training, receive future-depth supervision, and are then removed from the default inference graph.

% \paragraph{Efficient causal rollout.}
% Robot policies must balance prediction quality with action latency.
% Asynchronous WAM inference and training-only world modeling both suggest that the value of prediction can be separated from the cost of generating every future modality at deployment~\citep{lingbotva,fastwam,xwam}.
% \method{} adopts this separation explicitly: spatial registers learn from future depth targets during training, while default inference executes only the video-action main stream.

\section{Method}
\label{sec:method}

\begin{figure}[!t]
    \centering
    \includegraphics[page=1,width=1.0\linewidth]{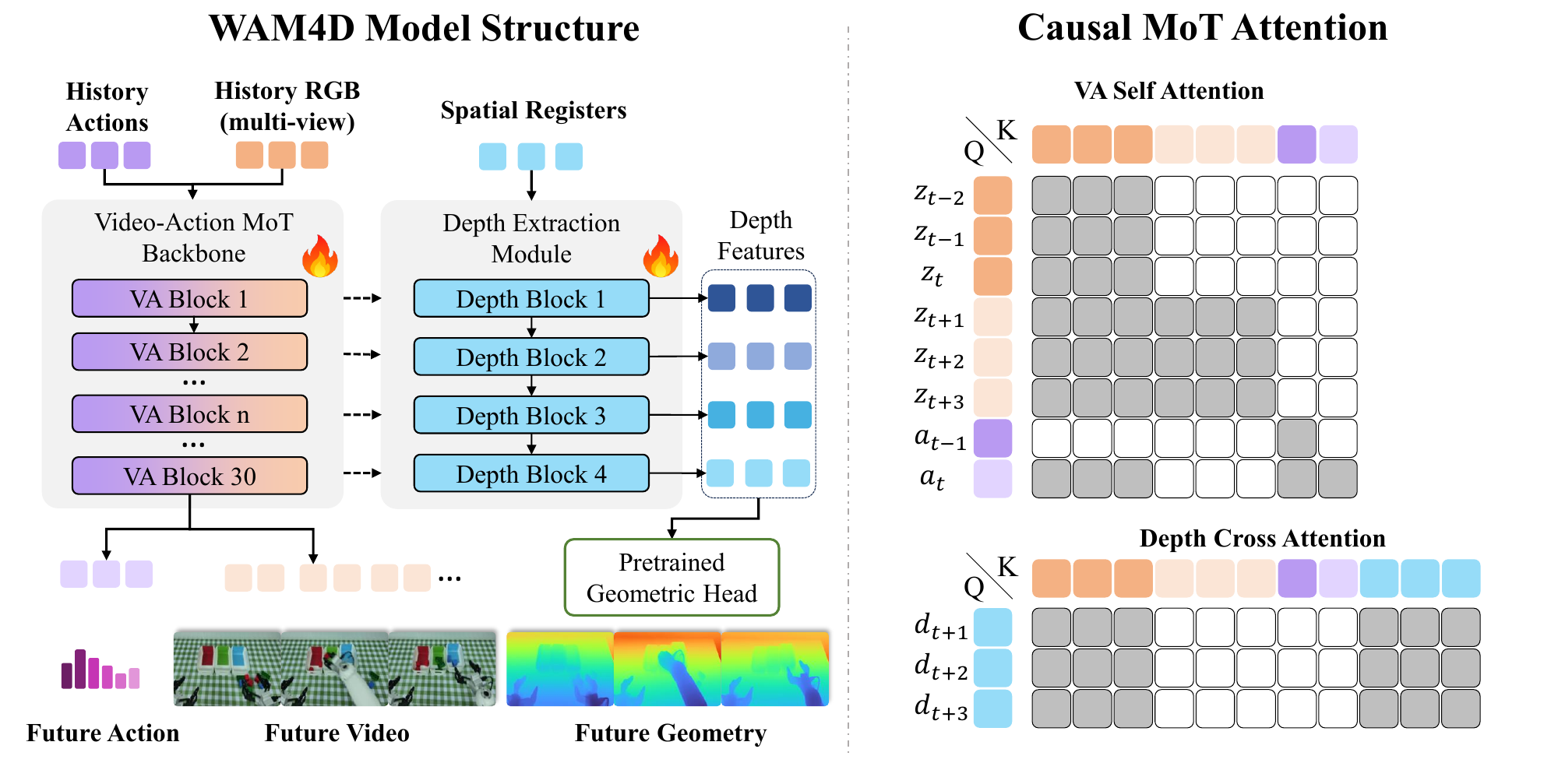}
    \caption{\method{} architecture and causal visibility pattern.}
    \label{fig:backbone}
\end{figure}

\subsection{WAM4D Backbone}

\method{} builds on the causal video-action backbone of LingBot-VA~\citep{lingbotva}.
At decision step $t$, the model conditions on a language instruction $l$, multi-view RGB history, and a queue of historical actions.
Let $O^{\mathrm{hist}}_t$ denote the history RGB mosaic sequence, $O^{\mathrm{fut}}_t$ the future RGB targets used during training, and $a_{i:j}$ the action sequence from step $i$ to $j$.
With historical action length $L_a$ and action prediction horizon $H_a$, the causal context is
\begin{equation}
    \mathcal{C}_t =
    \{l,\; O^{\mathrm{hist}}_t,\; a_{t-L_a:t-1}\},
\end{equation}
where future RGB frames and future actions are prediction targets rather than additional causal inputs.
The noised future video and action tokens are still included in the transformer sequence only as flow-matching states, so that the model can predict their corresponding flow targets.

For the video stream, a video VAE encodes the RGB sequence and splits the resulting latents into history tokens and clean future targets:
\begin{equation}
    [
    \mathbf{Z}^{\mathrm{hist}}_t,\;
    \mathbf{Z}^{\mathrm{fut}}_t
    ]
    =
    E_{\mathrm{vae}}
    \left(
    [
    O^{\mathrm{hist}}_t,\;
    O^{\mathrm{fut}}_t
    ]
    \right).
\end{equation}
During flow-matching training, the future video tokens fed to the backbone are noised states of these clean future targets, denoted as $\tilde{\mathbf{Z}}^{\mathrm{fut}}_t$.

The action stream is embedded in the same way:
\begin{equation}
    \mathbf{A}^{\mathrm{hist}}_t =
    E_a(a_{t-L_a:t-1}),
    \qquad
    \tilde{\mathbf{A}}^{\mathrm{fut}}_t =
    E_a(\tilde{a}_{t:t+H_a-1}),
\end{equation}
where $E_a$ is a lightweight action embedding layer, and $\tilde{a}_{t:t+H_a-1}$ denotes the flow-matching state of the future action chunk.

\method{} then follows causal WAMs~\citep{lingbotva} by jointly modeling future video latents and future actions in one video-action token sequence:
\begin{equation}
    \mathbf{X}^{(0)}_t =
    [
    \mathbf{Z}^{\mathrm{hist}}_t,\;
    \tilde{\mathbf{Z}}^{\mathrm{fut}}_t,\;
    \mathbf{A}^{\mathrm{hist}}_t,\;
    \tilde{\mathbf{A}}^{\mathrm{fut}}_t
    ].
\end{equation}
The sequence is processed by a video-action Mixture-of-Transformers (MoT) backbone:
\begin{equation}
    \mathbf{X}^{(\ell+1)}_t =
    \mathrm{VABlock}_{\ell}
    \left(
    \mathbf{X}^{(\ell)}_t;\mathbf{M}_{\mathrm{VA}}
    \right),
\end{equation}
where $\mathbf{M}_{\mathrm{VA}}$ is the causal visibility mask for the main video-action stream.
The video and action heads predict flow targets for the noised future video and action tokens, respectively.
The clean future video latents $\mathbf{Z}^{\mathrm{fut}}_t$ and the clean future action chunk $a_{t:t+H_a-1}$ are used to construct the corresponding training targets.

This main video-action path follows the causal WAM formulation, while our contribution lies in attaching a training-time geometry readout to its intermediate history video features.

\subsection{Spatial Register Distillation}

We introduce spatial register tokens as learnable geometry queries that extract future geometric information from causal video features.
Before being fed into the depth branch, a shared register grid is copied over the future depth timesteps and aligned with the multi-view RGB mosaic.
Since multi-camera observations are tiled into a single RGB canvas before VAE encoding, each register corresponds to a spatial region in the mosaic.

Let $\mathbf{R}_{\star}$ denote the learnable register grid, and let $\mathcal{T}_t$ denote the future depth supervision timesteps paired with the prediction targets.
The input registers are constructed as
\begin{equation}
    \mathbf{R}^{0}_t =
    \mathrm{Repeat}_{\tau \in \mathcal{T}_t}
    \left(
    \mathbf{R}_{\star}
    \right).
\end{equation}
Each copied register is associated with a target timestep and a mosaic pixel position.

Let $\mathbf{R}^{\ell}_t$ denote the register tokens at layer $\ell$, and let $\mathbf{Z}^{\mathrm{hist},\ell}_t$ denote the valid history video tokens from the video-action backbone at the same layer.
At selected layers $\ell \in \mathcal{L}_{r}$, the registers are updated by a depth extraction block.
The registers serve as queries, while the key-value set contains the registers themselves and history video tokens:
\begin{equation}
    \mathbf{R}^{\ell+1}_t
    =
    \mathrm{DepthBlock}_{\ell}
    \big(Q=\mathbf{R}^{\ell}_t,\; K,V=[\mathbf{R}^{\ell}_t,\mathbf{Z}^{\mathrm{hist},\ell}_t]\big).
\end{equation}
A standard RoPE encoding is applied inside the block using the target timestep and mosaic coordinates of the register and video tokens.
This update combines self-attention among registers with cross-attention from registers to history video features.

The updated register features are projected to the input space of a pretrained geometric head:
\begin{equation}
    \mathbf{G}_{t}
    =
    \mathcal{P}_{g}
    \left(\{\mathbf{R}^{\ell+1}_t\}_{\ell \in \mathcal{L}_{r}}\right),
    \qquad
    \hat{\mathbf{D}}^{\mathrm{fut}}_t
    =
    \mathcal{G}_{\phi}(\mathbf{G}_{t}).
\end{equation}
Here $\mathcal{P}_{g}$ is a lightweight projection layer, and $\mathcal{G}_{\phi}$ is the pretrained geometric head.
The output $\hat{\mathbf{D}}^{\mathrm{fut}}_t$ is the predicted future depth sequence.

The register branch is used only during training.
It predicts future depth from history video features, and the depth loss backpropagates into the shared video-action backbone.
By requiring future depth to be decoded from these register features, the pretrained geometric teacher distills its spatial priors into the history video features used by the backbone.

Our default readout places depth blocks after layers 12, 14, 16, and 18, and initializes the geometric head from Depth Anything V3~\citep{lin2025depth}.

\subsection{Causal Mixture Attention}

The model structure and causal visibility rules are summarized in Fig.~\ref{fig:backbone}. Text instructions are injected through cross attention and are visible to both video and action tokens. The main video-action stream follows a causal visibility pattern centered on future action prediction. At each denoising step, future action tokens can attend to history video tokens, history action tokens, and their own noised future action tokens. This gives the action predictor access to the causal observation and action context, while still allowing interaction among the action tokens being denoised.

To avoid non-causal shortcuts, future action tokens are masked from future video tokens and spatial registers. The future video tokens remain prediction targets for the video objective, rather than causal inputs for action generation. Spatial registers are also kept outside the policy path. They attend only to themselves and valid history video tokens, so the depth objective can shape causal video features without exposing auxiliary geometry tokens to the action predictor.

\begin{figure}[!t]
    \centering
    \includegraphics[page=1,width=1.0\linewidth]{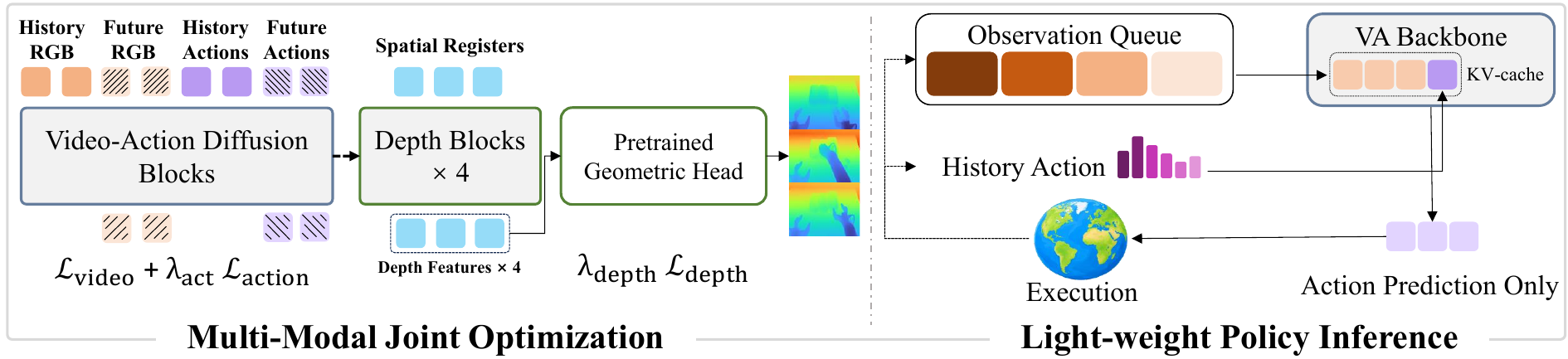}
    \caption{Training and inference paths of \method{}.}
    \label{fig:train_infer_differ}
\end{figure}

At inference, registers, depth blocks, and the geometric head are removed. The model reduces to a pure observation-to-action generation path. Training and inference paths are shown in Fig.~\ref{fig:train_infer_differ}.

\subsection{Training Objective}

The main video-action stream is trained with the conditional flow matching objective of the base causal WAM~\citep{lingbotva}.
We denote the future-video and future-action losses as $\mathcal{L}_{\mathrm{video}}$ and $\mathcal{L}_{\mathrm{action}}$, respectively.
For geometry distillation, future depth is supervised with a SmoothL1 loss.
Let $\mathcal{T}_t=\{t+1,\ldots,t+H_v\}$ be the set of future depth indices, where $H_v$ matches the future video horizon.
Let $\hat{D}_{\tau,p}$ and $D_{\tau,p}$ denote the predicted and target depth at time $\tau$ and pixel $p$.
Let $\Omega_\tau$ denote the set of valid depth pixels.
\begin{equation}
    \mathcal{L}_{\mathrm{depth}}
    =
    \frac{1}
    {\sum_{\tau \in \mathcal{T}_t} |\Omega_\tau|}
    \sum_{\tau \in \mathcal{T}_t}
    \sum_{p \in \Omega_\tau}
    \operatorname{SmoothL1}
    \left(
    \hat{D}_{\tau,p},
    D_{\tau,p}
    \right).
\end{equation}

The final objective is
\begin{equation}
    \mathcal{L}
    =
    \mathcal{L}_{\mathrm{video}}
    +
    \lambda_{\mathrm{act}}
    \mathcal{L}_{\mathrm{action}}
    +
    \lambda_{\mathrm{depth}}
    \mathcal{L}_{\mathrm{depth}}.
\end{equation}
We set $\lambda_{\mathrm{act}}=1$ and $\lambda_{\mathrm{depth}}=1$.
The depth loss updates the video-action backbone, spatial registers, depth blocks, projection layer, and geometric head.

\subsection{Implementation Details}
\label{app:implementation}

% \subsection{Default Configuration}

\noindent\textbf{Frame sampling strategy.}
The video-action backbone is initialized from the pretrained LingBot-VA base model and uses the Wan2.2 video VAE.
We sample at most 17 frames from each sequence.
A start frame id is randomly sampled over the sequence; when enough preceding frames are available, the history context contains 1, 5, or 9 frames.
After the start frame, 8 video-prediction target frames are sampled every 4 collected steps to match the VAE temporal stride.
If the remaining sequence is shorter than 8 target frames, the targets are padded to length 8.
For the latent video loss, only slots that can be encoded as valid VAE latents contribute to the loss, and latent slots containing padding frames are masked.
Depth targets use the same timestamps as the sampled video targets; unlike the latent-level video loss, depth supervision is frame-level, so every valid depth frame contributes to the depth loss.
The default action chunk size is 32.
For RoboTwin and AstriBot S1 experiments, the input is represented as a three-view mosaic consisting of one head camera and two wrist cameras.
The main view is resized to $256\times320$, and each wrist view is resized to $128\times160$.

\noindent\textbf{Action Prediction.}
The policy predicts only the absolute end-effector poses of the left and right arms.
Each arm is represented by a 3D position, a quaternion, and one gripper open/close value, giving a 16-dimensional action vector in total.
Position channels are normalized with percentile-based min--max normalization, i.e., quantile min--max normalization, using dataset-level $q_{01}$ and $q_{99}$ statistics.
Quaternion and gripper-opening channels use fixed lower and upper bounds of $-1$ and $1$.
All normalized action channels are represented in the range $[-1,1]$.

\noindent\textbf{View layout strategy.}
Spatial registers are aligned with the pixel layout of the mosaic.
Given the VAE spatial stride of 16 and the WAM transformer's $2\times2$ latent grouping, each register corresponds to a $32\times32$ cell in the input image.
The main view contributes an $8\times10$ register grid, each wrist view contributes a $4\times5$ register grid, and the tiled three-view mosaic forms a $12\times10$ register grid per future depth frame.
With 8 future-indexed depth frames, the default three-view model uses 960 spatial register tokens.
Unless stated otherwise, register cross-attention is applied after transformer layers 12, 14, 16, and 18.
Registers attend only to valid history video tokens and registers themselves, not to action tokens or future-video tokens.

\noindent\textbf{Geometric readout.}
The geometric readout uses a pretrained \texttt{DA3-GIANT-1.1} any-view DualDPT head.
Four learned linear adapters map WAM hidden states to the input dimension of the geometric head.

\noindent\textbf{Attention mask.}
The transformer sequence is partitioned into history-video tokens, future-video noise tokens, history-action tokens, future-action noise tokens, and register tokens.
The allowed self-attention pattern is:
\begin{center}
\small
\setlength{\tabcolsep}{5.0pt}
\renewcommand{\arraystretch}{1.02}
\begin{tabular}{ll}
\toprule
Query token group & Allowed key/value token groups \\
\midrule
Future action noise & History video, history action, future action noise \\
Register & Register, history video \\
Future video noise & History video, future video noise \\
History video & History video \\
History action & History action \\
\bottomrule
\end{tabular}
\end{center}

\noindent\textbf{Inference procedure.}
At deployment, the geometry path is removed and the policy maintains an observation queue and an executed-action history.
The inference loop is:
\begin{center}
\setlength{\fboxsep}{6pt}
\fbox{
\begin{minipage}{0.94\linewidth}
\small
\textbf{Algorithm 1: Deployment inference loop}\\[-0.2em]
\begin{tabular}{@{}r@{\hspace{0.8em}}p{0.82\linewidth}@{}}
1 & $Q_{\mathrm{obs}} \leftarrow \emptyset$, $A_{\mathrm{hist}} \leftarrow \emptyset$ \\
2 & \textbf{while} policy is running \textbf{do} \\
3 & \quad \textbf{if} $Q_{\mathrm{obs}}=\emptyset$ \textbf{then} capture the current observation and enqueue it into $Q_{\mathrm{obs}}$ \\
4 & \quad Encode $Q_{\mathrm{obs}}$ with the VAE to obtain video-context latents \\
5 & \quad Replace the video portion of the KV cache with the context latents \\
6 & \quad \textbf{if} $A_{\mathrm{hist}}\neq\emptyset$ \textbf{then} encode it and fill the action KV cache \\
7 & \quad Denoise future action tokens to obtain an action chunk \\
8 & \quad Execute the action chunk \\
9 & \quad During execution, capture one observation every 4 actions and enqueue it into $Q_{\mathrm{obs}}$ \\
10 & \quad $A_{\mathrm{hist}} \leftarrow$ executed actions \\
11 & \textbf{end while}
\end{tabular}
\end{minipage}
}
\end{center}

\noindent\textbf{Training Hyperparameters.}
Unless stated otherwise, the loss weights are set to $\lambda_{\mathrm{act}}=1$ and $\lambda_{\mathrm{depth}}=1$.
Training uses AdamW with learning rate $2\times10^{-5}\sqrt{N}$, where $N$ is the number of machines used for multi-machine parallel training, 10 warmup steps, gradient clipping at 2.0, bf16 parameters, a 50k-step budget for the main experiments, and a 10k-step budget for ablation experiments.

\section{Experiments}
\label{sec:experiments}

\subsection{Experimental Setup}
\label{app:datasets}
\noindent\textbf{Datasets and Tasks.}
We evaluate \method{} across simulation control, video and geometry quality, and real-world manipulation.

\noindent\textbf{RoboTwin 2.0.}
A unified policy is trained on the complete RoboTwin 2.0 task suite.
The clean setting comprises relatively uncluttered scenes, whereas the randomized setting introduces variation in clutter, illumination, background, tabletop height, object placement, and language instruction.
Geometry-supervised training uses re-collected RoboTwin demonstrations with depth annotations.
Each task contains 50 clean trajectories and 500 randomized trajectories.

\noindent\textbf{Real-world tasks.}
Real-world evaluation is conducted on the AstriBot S1 robot across four manipulation tasks: plate lifting, bottle placement, pen cap removal, and LEGO sorting.
The dataset contains 100 demonstrations per task, yielding 400 demonstrations in total.
Each method is evaluated with 10 physical rollouts per task.
Depth supervision for real-world demonstrations is likewise obtained with Depth Anything 3 following the same offline pseudo-depth pipeline.

\noindent\textbf{Baselines.}
\label{app:baselines}

The RoboTwin evaluation compares \method{} with VLA baselines ($\pi_0$, $\pi_{0.5}$) and recent video-action models, including Motus, LingBot-VA, and Fast-WAM.
Whenever possible, reproduced baselines use the same training and evaluation splits.
They also follow the same camera configuration and training budget.
For latency comparisons, all models use 10 action-denoising steps, and video-action models use 5 video-denoising steps when applicable.
The deployment configuration of \method{} removes the geometry path and performs action prediction only.

\noindent\textbf{Metrics.}
\label{app:metrics}

Policy performance is reported as the fraction of successful rollouts.
For real-world tasks, sub-action success records whether each task-specific intermediate goal is completed.

Video quality is evaluated with FVD, PSNR, SSIM, and LPIPS.
FVD measures the distribution gap between generated and ground-truth videos in a learned video-feature space; lower values indicate more realistic temporal generation.
PSNR measures pixel-level reconstruction fidelity, SSIM measures structural image similarity, and LPIPS measures perceptual distance in deep visual features.

Depth quality is evaluated with AbsRel and threshold accuracy $\delta_1$ and $\delta_2$ with respect to the available reference depth.
AbsRel is the mean relative absolute depth error, so lower is better.
$\delta_1$ and $\delta_2$ report the fraction of pixels whose predicted depth is within fixed multiplicative error thresholds of the target depth, so higher is better.

For point-cloud metrics, predicted and target depth maps are back-projected using a fixed dataset-level camera intrinsic matrix.
This protocol yields comparable point clouds for all methods.
CD$_1$ and CD$_2$ denote L1 and L2 Chamfer Distance variants between predicted and target point clouds.
F-score measures thresholded point-cloud overlap with the target geometry, while F-score-T measures the temporal consistency of this overlap across predicted future frames. 

% \noindent\textbf{Pseudo-depth label generation and reliability.}
% \label{app:pseudo_depth_reliability}

% RoboTwin depth targets use simulator ground-truth depth.
% For real-world demonstrations, the generated depth maps are used only as auxiliary training targets.
% They are never provided as observations to the policy and are removed together with the register branch at deployment.
% Therefore, pseudo-depth errors cannot serve as an additional inference-time sensing channel; they only act as a geometric regularizer on the causal visual representation.

% We process all frames offline using the frozen Depth Anything 3 (DA3) teacher.
% The predicted depth maps are resized back to the training resolution, converted to the same depth convention as simulator depth, and stored in uint16-mm format.
% Invalid or out-of-range pixels are excluded from the SmoothL1 depth loss.
% The same pseudo-depth generation pipeline is used for all methods that require depth supervision, ensuring that comparisons are not affected by different teacher labels.

\begin{table}[!tbp]
\centering
\begin{minipage}[t]{0.51\linewidth}
    \centering
    \caption{RoboTwin 2.0 full-task success rate. Action generation latency and VRAM are reported.}
    \label{tab:robotwin}
    \scriptsize
    \setlength{\tabcolsep}{2.0pt}
    \renewcommand{\arraystretch}{0.90}
    \resizebox{\linewidth}{!}{
    \begin{tabular}{lccccc}
        \toprule
        \multirow{2}{*}{Method} & \multirow{2}{*}{Clean} & \multirow{2}{*}{Rand.} & \multirow{2}{*}{Avg.} & Latency & VRAM \\
         &  &  &  &  (ms) & (GiB) \\
        \midrule
        $\pi_0$ & 65.9 & 58.4 & 62.2 & $64.16 \pm 0.06$ & \textbf{8.45} \\
        $\pi_{0.5}$ & 82.7 & 76.8 & 79.8 & $72.03 \pm 0.06$ & \textbf{8.45} \\
        Motus & 88.7 & 87.0 & 87.9 & $1516.30 \pm 10.64$ & 11.55 \\
        LingBot-VA & \underline{92.9} & \underline{91.6} & \textbf{92.3} & $843.57 \pm 11.55$ & 12.97 \\
        Fast-WAM & 91.9 & \textbf{91.8} & 91.8 & $425.53 \pm 6.01$ & 11.55  \\
        \midrule
        \method{} & \textbf{93.8} & 89.9 & \underline{91.8} & $525.43 \pm 5.64$ & \underline{9.71} \\
        \bottomrule
    \end{tabular}}
\end{minipage}
\hfill
\begin{minipage}[t]{0.48\linewidth}
    \centering
    \caption{Real-robot sub-action success over 10 rollouts per task.}
    \label{tab:real}
    \scriptsize
    \setlength{\tabcolsep}{1.5pt}
    \renewcommand{\arraystretch}{0.90}
    \resizebox{\linewidth}{!}{
    \begin{tabular}{@{}lcccccccc@{}}
        \toprule
        \multirow{2}{*}{Method} & Plate & Bottle & \multicolumn{3}{c}{Blocks} & \multicolumn{2}{c}{Pen} & \multirow{2}{*}{Avg.} \\
        \cmidrule(lr){2-2} \cmidrule(lr){3-3} \cmidrule(lr){4-6} \cmidrule(lr){7-8}
        & S1 & S1 & S1 & S2 & S3 & S1 & S2 & \\
        \midrule
        $\pi_{0.5}$ & 1.0 & 0.8 & 0.7 & 0.6 & 0.5 & 0.8 & 0.8 & 0.74 \\
        LingBot-VA & 1.0 & 1.0 & 1.0 & 0.7 & 0.4 & 0.9 & 0.9 & \underline{0.84} \\
        Fast-WAM & 0.9 & 1.0 & 0.8 & 0.7 & 0.5 & 0.9 & 0.8 & 0.80 \\
        \midrule
        \method{} & 0.9 & 0.9 & 1.0 & 0.9 & 0.8 & 0.9 & 0.9 & \textbf{0.90} \\
        \bottomrule
    \end{tabular}}
\end{minipage}
\end{table}

\begin{figure}[!tbp]
    \centering
    \includegraphics[width=1.0\linewidth]{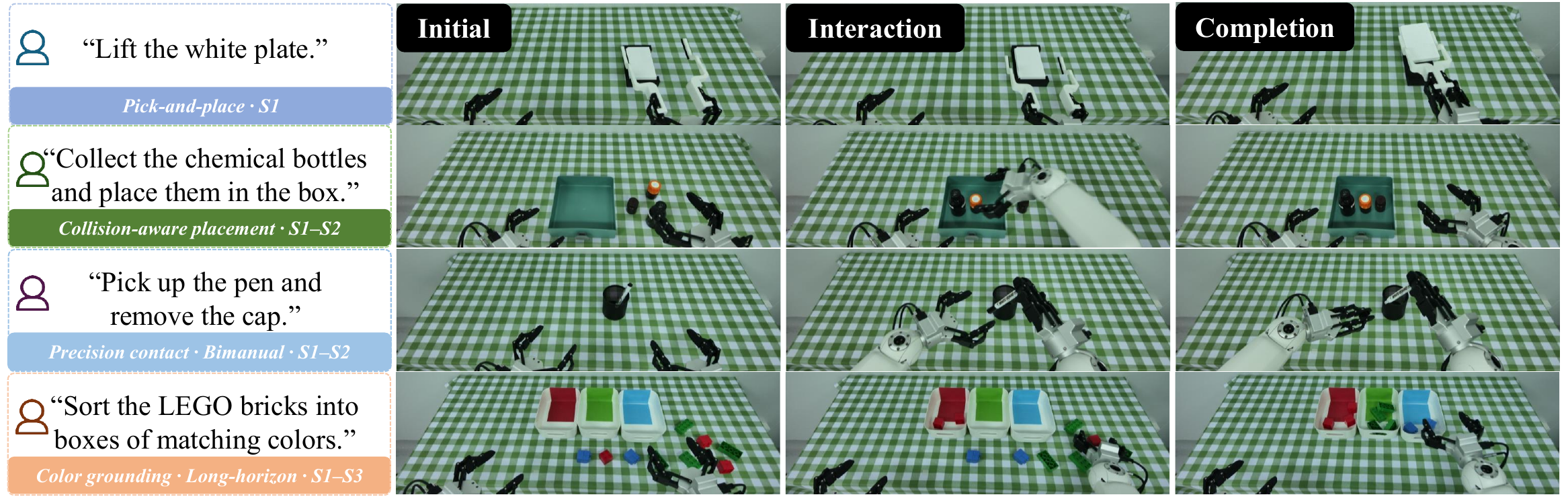}
    \caption{Real-world tasks on the AstriBot S1 platform. }
    \label{fig:real_tasks}
\end{figure}

\subsection{Simulation Results on RoboTwin 2.0}
% RoboTwin 2.0 provides a broad and challenging evaluation suite for robotic manipulation. It spans diverse task types, object configurations, scene variations, and bimanual control requirements.
We evaluate \method{} on the full RoboTwin 2.0 task suite and report the average success rate over 50 tasks under both clean and randomized settings. The full per-task success rates are reported in Tab.~\ref{tab:app_robotwin_full_results}.
We compare \method{} with strong VLA baselines and recent video-action models. The overall success rate is reported in Tab.~\ref{tab:robotwin}. \method{} achieves competitive performance while maintaining low inference cost. For a fair comparison, latency is measured under a unified inference setting with 10 action denoising steps and 5 video denoising steps for all applicable models. VRAM is reported as the peak allocated memory of the video-action backbone during inference.

\noindent\textbf{Full per-task results.}
We evaluate \method{} on the full RoboTwin 2.0 task suite and compare it with advanced baselines, including Fast-WAM, LingBot-VA, $\pi_0$, $\pi_{0.5}$, and Motus.
Table~\ref{tab:app_robotwin_full_results} reports the per-task success rates over 50 tasks under both Easy and Hard settings, corresponding to the clean and randomized evaluation protocols, respectively.
This full table complements the aggregate results in Tab.~\ref{tab:robotwin} and shows the task-level behavior of each method.

\begin{table}[!htbp]
    \centering
    \caption{Per-task RoboTwin 2.0 success rates. All entries are percentages. Easy and Hard correspond to the clean and randomized settings, respectively. Baseline results are taken from the full LingBot-VA RoboTwin 2.0 report~\citep{lingbotva} and the Fast-WAM paper~\citep{fastwam}.}
    \label{tab:app_robotwin_full_results}
    \scriptsize
    \setlength{\tabcolsep}{1.8pt}
    \renewcommand{\arraystretch}{0.92}
    \resizebox{\linewidth}{!}{
    \begin{tabular}{lccccccccccccc}
        \toprule
        Simulation Task & Horizon
        & \multicolumn{2}{c}{\method{}}
        & \multicolumn{2}{c}{Fast-WAM}
        & \multicolumn{2}{c}{LingBot-VA}
        & \multicolumn{2}{c}{$\pi_0$}
        & \multicolumn{2}{c}{$\pi_{0.5}$}
        & \multicolumn{2}{c}{Motus} \\
        \cmidrule(lr){3-4}
        \cmidrule(lr){5-6}
        \cmidrule(lr){7-8}
        \cmidrule(lr){9-10}
        \cmidrule(lr){11-12}
        \cmidrule(lr){13-14}
        & & Easy & Hard & Easy & Hard & Easy & Hard & Easy & Hard & Easy & Hard & Easy & Hard \\
        \midrule
        \textit{Adjust Bottle} & 1 & \textbf{100} & \textbf{99} & 100 & 100 & 90 & 94 & 99 & 95 & 100 & 99 & 89 & 93 \\
        \textit{Beat Block Hammer} & 1 & \textbf{99} & 97 & 99 & 97 & 96 & 98 & 79 & 84 & 96 & 93 & 95 & 88 \\
        \textit{Blocks Ranking RGB} & 3 & \textbf{100} & 97 & 100 & 100 & 99 & 98 & 80 & 63 & 92 & 85 & 99 & 97 \\
        \textit{Blocks Ranking Size} & 3 & \textbf{98} & 84 & 94 & 98 & 94 & 96 & 14 & 5 & 49 & 26 & 75 & 63 \\
        \textit{Click Alarmclock} & 1 & \textbf{100} & \textbf{100} & 100 & 100 & 99 & 100 & 77 & 68 & 98 & 89 & 100 & 100 \\
        \textit{Click Bell} & 1 & \textbf{100} & \textbf{100} & 100 & 100 & 100 & 100 & 71 & 48 & 99 & 66 & 100 & 100 \\
        \textit{Dump Bin Bigbin} & 1 & 92 & 94 & 97 & 96 & 89 & 96 & 88 & 83 & 92 & 97 & 95 & 91 \\
        \textit{Grab Roller} & 1 & \textbf{100} & \textbf{100} & 100 & 100 & 100 & 100 & 98 & 94 & 100 & 100 & 100 & 100 \\
        \textit{Handover Block} & 2 & 96 & \textbf{93} & 95 & 81 & 99 & 78 & 47 & 31 & 66 & 57 & 86 & 73 \\
        \textit{Handover Mic} & 2 & 94 & 94 & 99 & 100 & 94 & 96 & 97 & 97 & 98 & 97 & 78 & 63 \\
        \textit{Hanging Mug} & 2 & \textbf{64} & \textbf{57} & 58 & 62 & 40 & 28 & 14 & 11 & 18 & 17 & 38 & 39 \\
        \textit{Lift Pot} & 1 & \textbf{100} & 88 & 100 & 100 & 100 & 99 & 80 & 72 & 96 & 85 & 96 & 99 \\
        \textit{Move Can Pot} & 1 & \textbf{96} & \textbf{97} & 90 & 88 & 94 & 97 & 68 & 48 & 51 & 55 & 34 & 74 \\
        \textit{Move Pillbottle Pad} & 1 & \textbf{100} & \textbf{99} & 100 & 99 & 99 & 99 & 67 & 46 & 84 & 61 & 93 & 96 \\
        \textit{Move Playingcard Away} & 1 & \textbf{100} & \textbf{100} & 100 & 100 & 100 & 99 & 74 & 65 & 96 & 84 & 100 & 96 \\
        \textit{Move Stapler Pad} & 1 & 79 & 76 & 77 & 64 & 91 & 79 & 41 & 24 & 56 & 42 & 83 & 85 \\
        \textit{Open Laptop} & 1 & \textbf{100} & 66 & 98 & 100 & 92 & 94 & 71 & 81 & 90 & 96 & 95 & 91 \\
        \textit{Open Microwave} & 1 & 61 & 65 & 62 & 45 & 82 & 86 & 4 & 32 & 34 & 77 & 95 & 91 \\
        \textit{Pick Diverse Bottles} & 2 & \textbf{98} & \textbf{92} & 80 & 85 & 89 & 82 & 69 & 31 & 81 & 71 & 90 & 91 \\
        \textit{Pick Dual Bottles} & 2 & \textbf{100} & \textbf{99} & 100 & 96 & 100 & 99 & 59 & 37 & 93 & 63 & 96 & 90 \\
        \textit{Place A2B Left} & 1 & \textbf{97} & 86 & 95 & 93 & 97 & 93 & 43 & 47 & 87 & 82 & 82 & 79 \\
        \textit{Place A2B Right} & 1 & 94 & 79 & 93 & 99 & 97 & 95 & 39 & 34 & 87 & 84 & 90 & 87 \\
        \textit{Place Bread Basket} & 1 & 93 & \textbf{97} & 91 & 93 & 97 & 95 & 62 & 46 & 77 & 64 & 91 & 94 \\
        \textit{Place Bread Skillet} & 2 & 93 & 87 & 90 & 93 & 95 & 90 & 66 & 49 & 85 & 66 & 86 & 83 \\
        \textit{Place Burger Fries} & 2 & 96 & 97 & 96 & 99 & 97 & 95 & 81 & 76 & 94 & 87 & 98 & 98 \\
        \textit{Place Can Basket} & 2 & \textbf{87} & 83 & 71 & 69 & 81 & 84 & 55 & 46 & 62 & 62 & 81 & 76 \\
        \textit{Place Cans Plasticbox} & 2 & \textbf{100} & \textbf{100} & 99 & 96 & 100 & 99 & 63 & 45 & 94 & 84 & 98 & 94 \\
        \textit{Place Container Plate} & 1 & 98 & 91 & 96 & 100 & 99 & 97 & 97 & 92 & 99 & 95 & 98 & 99 \\
        \textit{Place Dual Shoes} & 2 & \textbf{96} & \textbf{97} & 94 & 88 & 94 & 89 & 59 & 51 & 75 & 75 & 93 & 87 \\
        \textit{Place Empty Cup} & 1 & \textbf{100} & \textbf{100} & 100 & 100 & 100 & 100 & 91 & 85 & 100 & 99 & 99 & 98 \\
        \textit{Place Fan} & 1 & \textbf{100} & \textbf{95} & 96 & 96 & 99 & 93 & 66 & 71 & 87 & 85 & 91 & 87 \\
        \textit{Place Mouse Pad} & 1 & \textbf{96} & \textbf{97} & 83 & 89 & 93 & 96 & 20 & 20 & 60 & 39 & 66 & 68 \\
        \textit{Place Object Basket} & 2 & 88 & \textbf{90} & 89 & 88 & 91 & 88 & 67 & 70 & 80 & 76 & 81 & 87 \\
        \textit{Place Object Scale} & 1 & \textbf{99} & \textbf{97} & 90 & 97 & 96 & 95 & 57 & 52 & 86 & 80 & 88 & 85 \\
        \textit{Place Object Stand} & 1 & 98 & \textbf{100} & 90 & 94 & 99 & 96 & 82 & 68 & 91 & 85 & 98 & 97 \\
        \textit{Place Phone Stand} & 1 & 94 & 81 & 97 & 99 & 97 & 97 & 49 & 53 & 81 & 81 & 87 & 86 \\
        \textit{Place Shoe} & 1 & \textbf{100} & \textbf{99} & 96 & 99 & 98 & 98 & 76 & 76 & 92 & 93 & 99 & 97 \\
        \textit{Press Stapler} & 1 & \textbf{97} & 94 & 90 & 97 & 85 & 82 & 44 & 37 & 87 & 83 & 93 & 98 \\
        \textit{Put Bottles Dustbin} & 3 & 87 & 85 & 95 & 90 & 87 & 91 & 65 & 56 & 84 & 79 & 81 & 79 \\
        \textit{Put Object Cabinet} & 2 & 81 & 53 & 94 & 89 & 85 & 87 & 73 & 60 & 80 & 79 & 88 & 71 \\
        \textit{Rotate QRcode} & 1 & 95 & \textbf{95} & 93 & 89 & 96 & 91 & 74 & 70 & 89 & 87 & 89 & 73 \\
        \textit{Scan Object} & 2 & 92 & 68 & 89 & 92 & 96 & 91 & 55 & 42 & 72 & 65 & 67 & 66 \\
        \textit{Shake Bottle Horizontally} & 1 & \textbf{100} & 99 & 100 & 100 & 100 & 99 & 98 & 92 & 99 & 99 & 100 & 98 \\
        \textit{Shake Bottle} & 1 & \textbf{100} & 99 & 100 & 100 & 100 & 97 & 94 & 91 & 99 & 97 & 100 & 97 \\
        \textit{Stack Blocks Three} & 3 & 98 & 97 & 95 & 97 & 99 & 98 & 72 & 52 & 91 & 76 & 91 & 95 \\
        \textit{Stack Blocks Two} & 2 & \textbf{100} & \textbf{100} & 100 & 100 & 100 & 98 & 93 & 79 & 97 & 100 & 100 & 98 \\
        \textit{Stack Bowls Three} & 3 & 83 & 84 & 80 & 81 & 86 & 83 & 77 & 75 & 77 & 71 & 79 & 87 \\
        \textit{Stack Bowls Two} & 2 & 96 & 95 & 92 & 98 & 94 & 98 & 94 & 95 & 95 & 96 & 98 & 98 \\
        \textit{Stamp Seal} & 1 & \textbf{96} & 87 & 90 & 94 & 96 & 97 & 46 & 33 & 79 & 55 & 93 & 92 \\
        \textit{Turn Switch} & 1 & 60 & 64 & 61 & 59 & 44 & 45 & 41 & 42 & 62 & 54 & 84 & 78 \\
        \midrule
        Average (50 tasks) & -- & 93.82 & 89.86 & 91.88 & 91.78 & 92.90 & 91.50 & 65.92 & 58.40 & 82.74 & 76.76 & 88.52 & 87.04 \\
        \bottomrule
    \end{tabular}
    }
\end{table}

\subsection{Real-World Results}
We evaluate \method{} on the AstriBot S1 platform using four different manipulation scenarios. Fig.~\ref{fig:real_tasks} illustrates the evaluation setup, and the sub-action definitions are reported in Tab.~\ref{tab:app_real_subactions}. The quantitative results are reported in Tab.~\ref{tab:real}. \method{} achieves the best overall performance across all four real-robot evaluations, demonstrating its robustness in contact-rich, geometry-sensitive, and long-horizon manipulation.
LingBot-VA shows lower success rates in these settings. In its default code implementation, action predictions are anchored to the initial action history, which can cause global trajectory drift when the initial configuration changes. Its long history window may further introduce stale action information that conflicts with the current observation, reducing reliability during precise contact and long-horizon execution. In our tests, $\pi_{0.5}$ and Fast-WAM show a narrower effective workspace, failing more often when object positions move farther from the nominal region.

\noindent\textbf{Sub-action definitions.}
Table~\ref{tab:app_real_subactions} defines the sub-actions used for real-robot evaluation on the AstriBot S1 platform.
During each rollout, a sub-action is counted as successful when the robot completes the corresponding intermediate goal.
Because the tasks are sequential, if any sub-action fails, subsequent sub-actions are not attempted and are recorded as 0.

\begin{table}[!htbp]
    \centering
    \caption{Sub-action definitions for the AstriBot S1 real-world tasks.}
    \label{tab:app_real_subactions}
    \small
    \setlength{\tabcolsep}{4.0pt}
    \renewcommand{\arraystretch}{1.05}
    \begin{tabular}{lll}
        \toprule
        Task & Step & Success criterion \\
        \midrule
        Plate & S1 & Grasp and lift the white plate. \\
        \multirow{2}{*}{Bottles} & S1 & Grasp a chemical bottle. \\
         & S2 & Place the grasped bottle into the tray. \\
        \multirow{3}{*}{Lego} & S1 & Place the first colored Lego block into its matching box. \\
         & S2 & Place the second colored Lego block into its matching box. \\
         & S3 & Place the third colored Lego block into its matching box. \\
        \multirow{2}{*}{Pen} & S1 & Grasp the pen. \\
         & S2 & Remove the pen cap from the pen. \\
        \bottomrule
    \end{tabular}
\end{table}

\FloatBarrier

\begin{figure}[!t]
    \centering
    \includegraphics[page=1,width=0.98\linewidth]{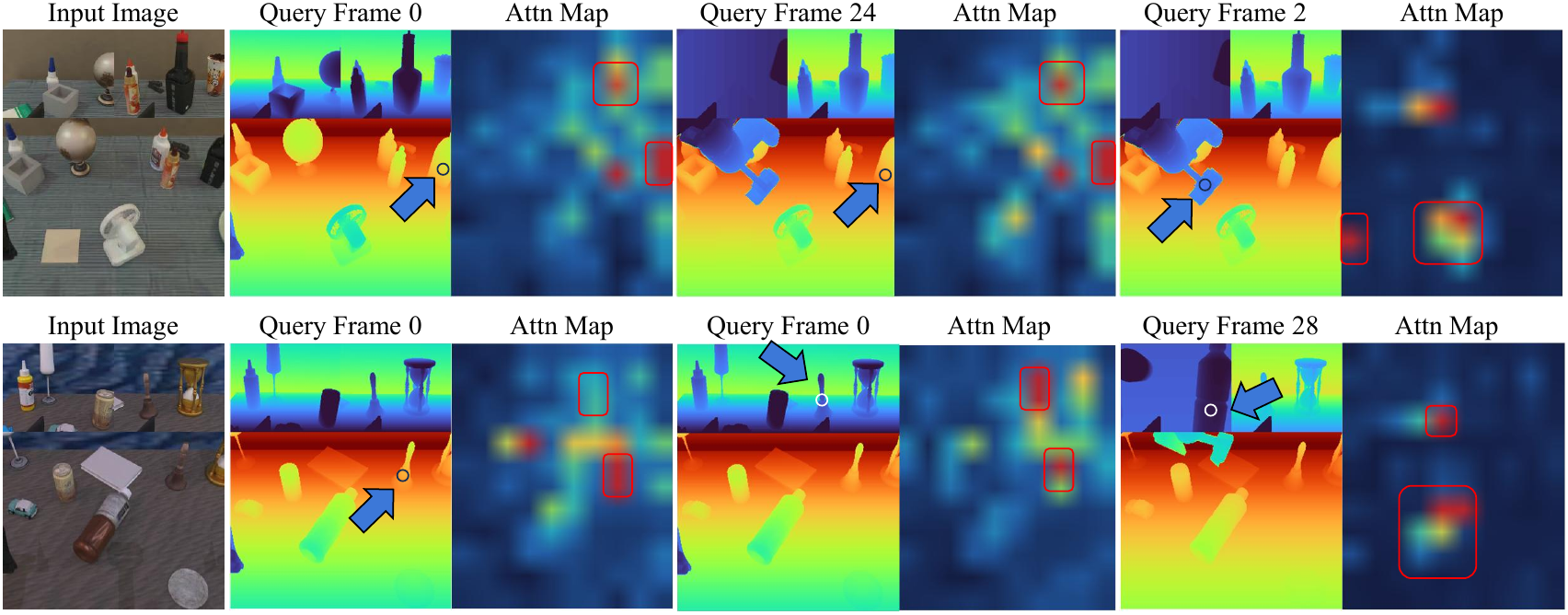}
    \caption{Spatial register attention on RoboTwin randomized samples.}
    \label{fig:attn_vis}
\end{figure}

\subsection{Model Analysis and Ablations}
All ablations use the same data, backbone size, history length, action horizon, optimizer, and training budget unless stated otherwise.
Each comparison controls all variables except the component under study.
% We train each ablation on an 8-GPU setup with batch size 8 per GPU for 10{,}000 optimizer steps.
% We report both control metrics and geometry metrics.
We therefore evaluate both generation quality and control performance, since a lower depth loss alone does not show that geometry improves manipulation.
The ten-task split and full ablation metrics are reported in Tab.~\ref{tab:app_robotwin_10task_split} and Tab.~\ref{tab:app_robotwin_10task_counts}.

\noindent\textbf{Ablation split and full metrics.}
\begin{table}[!htbp]
    \centering
    \caption{RoboTwin ten-task split used for model-analysis ablations.}
    \label{tab:app_robotwin_10task_split}
    \small
    \setlength{\tabcolsep}{5.0pt}
    \renewcommand{\arraystretch}{1.02}
    \begin{tabular}{lc}
        \toprule
        Simulation Task & Horizon \\
        \midrule
        \textit{Handover Block} & 2 \\
        \textit{Hanging Mug} & 2 \\
        \textit{Move Stapler Pad} & 1 \\
        \textit{Open Microwave} & 1 \\
        \textit{Pick Diverse Bottles} & 2 \\
        \textit{Place Can Basket} & 2 \\
        \textit{Press Stapler} & 1 \\
        \textit{Put Object Cabinet} & 2 \\
        \textit{Stack Bowls Three} & 3 \\
        \textit{Turn Switch} & 1 \\
        \bottomrule
    \end{tabular}
\end{table}

\begin{table}[!htbp]
    \centering
    \caption{Per-task success rates and full ablation metrics for the RoboTwin ten-task runs.}
    \label{tab:app_robotwin_10task_counts}
    \scriptsize
    \setlength{\tabcolsep}{2.2pt}
    \renewcommand{\arraystretch}{0.92}
    \resizebox{\linewidth}{!}{
    \begin{tabular}{lccccccccc}
        \toprule
        Metric / Task & No Depth & Bi-dir & Deep & Middle & Shallow & Uniform & VAE DH & Rand. DA3 & Train. DA3 \\
        \midrule
        Transformer Params & 5.089B & 5.841B & 5.690B & 5.690B & 5.690B & 5.690B & 5.744B & 5.690B & 5.690B \\
        Clean SR (100-eval tasks) & 71.7\% & \underline{76.6\%} & 74.5\% & 75.2\% & 72.5\% & 70.6\% & 70.7\% & 70.0\% & \textbf{80.1\%} \\
        Randomized SR (100-eval tasks) & 69.1\% & 72.5\% & 67.4\% & 69.7\% & 66.2\% & 67.0\% & 68.6\% & \underline{74.8\%} & \textbf{75.4\%} \\
        \midrule
        \multicolumn{10}{l}{\textbf{Video metrics}} \\
        FVD $\downarrow$ & 181.2 & 175.3 & 171.5 & 179.8 & \underline{168.8} & 175.3 & 178.0 & 189.8 & \textbf{164.5} \\
        PSNR $\uparrow$ & 20.39 & 20.75 & 20.63 & 20.67 & \textbf{21.19} & 20.57 & 20.72 & 20.22 & \underline{21.13} \\
        SSIM $\uparrow$ & 0.795 & 0.798 & 0.797 & 0.798 & \underline{0.808} & 0.798 & 0.802 & 0.794 & \textbf{0.811} \\
        LPIPS $\downarrow$ & 0.165 & 0.155 & 0.158 & 0.159 & \textbf{0.143} & 0.162 & \underline{0.153} & 0.162 & \textbf{0.143} \\
        \midrule
        \multicolumn{10}{l}{\textbf{Depth and point-cloud metrics}} \\
        $\delta_1 \uparrow$ & \multirow{7}{*}{--} & 0.916 & 0.931 & \underline{0.945} & 0.940 & 0.941 & 0.901 & 0.941 & \textbf{0.948} \\
        $\delta_2 \uparrow$ & & 0.952 & 0.963 & 0.972 & 0.971 & 0.971 & 0.950 & \underline{0.973} & \textbf{0.974} \\
        AbsRel $\downarrow$ & & 0.074 & 0.064 & \underline{0.053} & 0.058 & 0.058 & 0.081 & 0.059 & \textbf{0.049} \\
        CD$_1 \downarrow$ & & 0.0151 & 0.0130 & \underline{0.0108} & 0.0128 & 0.0121 & 0.0171 & 0.0129 & \textbf{0.0099} \\
        CD$_2 \downarrow$ & & 0.0007 & 0.0005 & \underline{0.0004} & 0.0005 & 0.0005 & 0.0009 & 0.0005 & \textbf{0.0003} \\
        F-score $\uparrow$ & & 0.579 & 0.621 & \underline{0.685} & 0.613 & 0.652 & 0.519 & 0.589 & \textbf{0.710} \\
        F-score-T $\uparrow$ & & 0.769 & 0.776 & \underline{0.825} & 0.784 & 0.824 & 0.685 & 0.737 & \textbf{0.848} \\
        \midrule
        \multicolumn{10}{l}{\textbf{Clean}} \\
        \textit{Handover Block} & 97/100 & 98/100 & \textbf{100/100} & 98/100 & \textbf{100/100} & 98/100 & \underline{99/100} & 78/100 & \textbf{100/100} \\
        \textit{Hanging Mug} & 58/100 & 57/100 & \underline{82/100} & 79/100 & 56/100 & 53/100 & 60/100 & 45/100 & \textbf{85/100} \\
        \textit{Move Stapler Pad} & 66/100 & \underline{69/100} & \underline{69/100} & 67/100 & \textbf{76/100} & 56/100 & 51/100 & 49/100 & 65/100 \\
        \textit{Open Microwave} & 39/100 & \underline{69/100} & 45/100 & 53/100 & 44/100 & 50/100 & 40/100 & 53/100 & \textbf{78/100} \\
        \textit{Pick Diverse Bottles} & \textbf{94/100} & 92/100 & 91/100 & 90/100 & 91/100 & 91/100 & \underline{93/100} & \underline{93/100} & \textbf{94/100} \\
        \textit{Place Can Basket} & 68/100 & 72/100 & 67/100 & \textbf{81/100} & 64/100 & 60/100 & \underline{77/100} & 69/100 & 75/100 \\
        \textit{Press Stapler} & 95/100 & 94/100 & 94/100 & 96/100 & 93/100 & 95/100 & \textbf{99/100} & 92/100 & \underline{97/100} \\
        \textit{Put Object Cabinet} & 54/100 & 52/100 & 55/100 & 46/100 & 47/100 & 49/100 & 48/100 & \underline{66/100} & \textbf{87/100} \\
        \textit{Stack Bowls Three} & 78/100 & \underline{85/100} & 77/100 & 81/100 & \textbf{88/100} & 81/100 & 79/100 & 70/100 & 74/100 \\
        \textit{Turn Switch} & 68/100 & \underline{78/100} & 65/100 & 61/100 & 66/100 & 73/100 & 61/100 & \textbf{85/100} & 46/100 \\
        Average (Clean) & 71.7\% & \underline{76.6\%} & 74.5\% & 75.2\% & 72.5\% & 70.6\% & 70.7\% & 70.0\% & \textbf{80.1\%} \\
        \midrule
        \multicolumn{10}{l}{\textbf{Randomized}} \\
        \textit{Handover Block} & 75/100 & \underline{95/100} & 79/100 & 81/100 & 70/100 & 77/100 & 72/100 & \textbf{97/100} & \textbf{97/100} \\
        \textit{Hanging Mug} & 66/100 & 69/100 & 72/100 & 64/100 & 66/100 & 71/100 & 72/100 & \underline{79/100} & \textbf{81/100} \\
        \textit{Move Stapler Pad} & \textbf{61/100} & 46/100 & 49/100 & 55/100 & \underline{58/100} & 54/100 & 57/100 & 49/100 & 49/100 \\
        \textit{Open Microwave} & 41/100 & 78/100 & 34/100 & 53/100 & 35/100 & 31/100 & 44/100 & \textbf{80/100} & \underline{79/100} \\
        \textit{Pick Diverse Bottles} & 80/100 & 89/100 & \textbf{91/100} & 87/100 & 80/100 & 88/100 & 80/100 & \underline{90/100} & 89/100 \\
        \textit{Place Can Basket} & \textbf{75/100} & 72/100 & 72/100 & 68/100 & \underline{74/100} & 71/100 & 73/100 & 72/100 & \textbf{75/100} \\
        \textit{Press Stapler} & 94/100 & 95/100 & 93/100 & 95/100 & 94/100 & 94/100 & 95/100 & \underline{97/100} & \textbf{98/100} \\
        \textit{Put Object Cabinet} & 43/100 & 47/100 & 47/100 & 48/100 & 45/100 & 45/100 & \textbf{51/100} & \underline{49/100} & 37/100 \\
        \textit{Stack Bowls Three} & \textbf{88/100} & 75/100 & 85/100 & 82/100 & 75/100 & 80/100 & 81/100 & 79/100 & \underline{87/100} \\
        \textit{Turn Switch} & \textbf{68/100} & 59/100 & 52/100 & 64/100 & \underline{65/100} & 59/100 & 61/100 & 56/100 & 62/100 \\
        Average (Randomized) & 69.1\% & 72.5\% & 67.4\% & 69.7\% & 66.2\% & 67.0\% & 68.6\% & \underline{74.8\%} & \textbf{75.4\%} \\
        \bottomrule
    \end{tabular}
    }
\end{table}

Model-analysis ablations are conducted on a fixed ten-task subset of RoboTwin 2.0 to maintain computational tractability while preserving diverse manipulation requirements.
The split covers single-step and multi-step tasks involving hanging, articulated-object interaction, contact pressing, placement, handover, object selection, and stacking.
Table~\ref{tab:app_robotwin_10task_split} lists the tasks and horizons used by all ablation runs.
Table~\ref{tab:app_robotwin_10task_counts} consolidates per-task success rates, model size, and full video/geometry metrics for the nine ten-task ablation runs.
No Depth is an otherwise identical model trained under the same conditions without any depth branch or depth loss.
VAE DH discretizes depth into 0--255 values and uses the VAE as a depth head to decode depth latents; its depth readout is taken from the output of the layer-26 VA block.
All remaining variants use Spatial Registers as the depth readout interface.
For the layer configurations, Shallow, Middle, Deep, and Uniform registers insert cross-attention after layers 2/4/6/8, 12/14/16/18, 22/24/26/28, and 6/12/18/24, respectively; Bi-dir uses layers 6/12/18/24 with bidirectional visibility.
Rand. DA3 and Train. DA3 use the default middle-layer Spatial Register interface with the DA3 head trained from random and pretrained initialization, respectively.
Within each summary or per-task row, the best entry is bolded and the second-best entry is underlined, with ties marked together.

\subsubsection{Spatial Register Attention Analysis}
We visualize register-to-history attention on RoboTwin randomized samples in Fig.~\ref{fig:attn_vis}. The attention maps are extracted from selected transformer layers and averaged over heads. They show which history image regions are queried by spatial registers when predicting future depth. The visualizations show that spatial registers capture meaningful geometric cues from the history context. Object-related registers often attend to the same object across views, while background-related registers focus on geometrically consistent static regions. Gripper-related registers show strong attention around the initial gripper pose, suggesting that the model uses early action-history cues to predict future geometry.

\subsubsection{Spatial Register Design Ablation}

\begin{table}[!t]
    \centering
    \caption{Ablation of depth readout interface, register placement, and register visibility on the RoboTwin 10-task split with the depth head fixed. Selected quality metrics are reported.}
    \label{tab:ablation_register_depth}
    \scriptsize
    \setlength{\tabcolsep}{2.0pt}
    \renewcommand{\arraystretch}{0.92}
    \resizebox{\linewidth}{!}{
    \begin{tabular}{llcccccccccc}
        \toprule
        Variant & Depth Readout & Layers & Clean SR & FVD $\downarrow$ & PSNR $\uparrow$ & LPIPS $\downarrow$ & AbsRel $\downarrow$ & $\delta_1 \uparrow$ & CD$_1 \downarrow$ & F-score $\uparrow$ & F-score-T $\uparrow$ \\
        \midrule
        No depth & None & -- & 71.7 & 181.2 & 20.39 & 0.165 & \multicolumn{5}{c}{--} \\
        VAE depth head & Future video hiddens & 27, 28, 29, 30 & 70.7 & 178.0 & 20.72 & \underline{0.153} & 0.081 & 0.901 & 0.0171 & 0.519 & 0.685 \\
        \midrule
        Shallow registers & \multirow{5}{*}{Spatial registers} & 2, 4, 6, 8 & 72.5 & \textbf{168.8} & \textbf{21.19} & \textbf{0.143} & 0.058 & 0.940 & 0.0128 & 0.613 & 0.784 \\
        Middle registers & & 12, 14, 16, 18 & \underline{75.2} & 179.8 & 20.67 & 0.159 & \textbf{0.053} & \textbf{0.945} & \textbf{0.0108} & \textbf{0.685} & \textbf{0.825} \\
        Deep registers & & 22, 24, 26, 28 & 74.5 & \underline{171.5} & 20.63 & 0.158 & 0.064 & 0.931 & 0.0130 & 0.621 & 0.776 \\
        Uniform registers & & 6, 12, 18, 24 & 70.6 & 175.3 & 20.57 & 0.162 & \underline{0.058} & \underline{0.941} & \underline{0.0121} & \underline{0.652} & \underline{0.824} \\
        Bidirectional registers & & 6, 12, 18, 24 & \textbf{76.6} & 175.3 & \underline{20.75} & 0.155 & 0.074 & 0.916 & 0.0151 & 0.579 & 0.769 \\
        \bottomrule
    \end{tabular}}
\end{table}

Tab.~\ref{tab:ablation_register_depth} studies three design choices: the depth readout interface, the register insertion layers, and the visibility between registers and the main video-action stream. Adding a direct depth head to future VAE latents improves RGB generation over the no-depth baseline, but its geometry metrics remain weaker than the spatial-register variants. This is likely because RGB VAE latents are not designed for accurate metric depth encoding and decoding without retraining the VAE. In contrast, spatial registers query only history video tokens, so the depth objective directly shapes the history video features used by action prediction. The pretrained geometric head also provides a more precise depth readout than a lightweight VAE-latent decoder.

Register placement reveals different trade-offs between visual synthesis and geometric distillation. Shallow registers achieve the best RGB metrics, possibly because early geometry regularization encourages the denoising backbone to extract useful structure from noisy features, which benefits the overall denoising process. However, they provide weaker control and geometry quality than middle-layer registers. Middle-layer registers achieve the best balance between geometric distillation and visual feature preservation, leading to the strongest unidirectional control performance and the best overall geometry quality. Deep and uniform placements are less effective, likely because late features are more specialized for denoising output prediction, while uniform insertion spreads the geometric readout across layers with different abstraction levels.

The bidirectional variant obtains the highest success rate, but it requires the main video-action stream to read register features, introducing additional computation and model complexity. It also degrades most geometry metrics compared with the middle-layer unidirectional setting. Therefore, we choose unidirectional registers at layers 12, 14, 16, and 18 as the default design. This setting balances geometric distillation with visual feature preservation, while also providing a practical trade-off between control performance and training cost.

\subsubsection{Geometric Head Ablation}
\begin{table}[!t]
    \centering
    \caption{Geometric head ablations on the RoboTwin 10-task split. }
    \label{tab:ablation_teacher}
    \scriptsize
    \setlength{\tabcolsep}{2.0pt}
    \renewcommand{\arraystretch}{0.92}
    \resizebox{\linewidth}{!}{
    \begin{tabular}{lccccccccc}
        \toprule
        Variant & Clean SR & FVD $\downarrow$ & PSNR $\uparrow$ & LPIPS $\downarrow$ & AbsRel $\downarrow$ & $\delta_1 \uparrow$ & CD$_1 \downarrow$ & F-score $\uparrow$ & F-score-T $\uparrow$ \\
        \midrule
        w/o depth & 71.7 & 181.2 & 20.39 & 0.165 & \multicolumn{5}{c}{--} \\
        Trainable random init & 70.0 & 189.8 & 20.22 & 0.162 & 0.059 & 0.941 & 0.0129 & 0.589 & 0.737 \\
        Trainable pretrained & \textbf{80.1} & \textbf{164.5} & \textbf{21.13} & \textbf{0.143} & \textbf{0.049} & \textbf{0.948} & \textbf{0.0099} & \textbf{0.710} & \textbf{0.848} \\
        Fixed pretrained & \underline{75.2} & \underline{179.8} & \underline{20.67} & \underline{0.159} & \underline{0.053} & \underline{0.945} & \underline{0.0108} & \underline{0.685} & \underline{0.825} \\
        \bottomrule
    \end{tabular}
    }
\end{table}

We compare different depth supervision paths with the same register interface.
A randomly initialized depth head tests whether ordinary depth prediction is sufficient.
A tuned pretrained geometric head tests whether adaptation is needed.
A fixed pretrained geometric head tests whether geometric foundation knowledge is sufficient without adaptation.

Tab.~\ref{tab:ablation_teacher} shows that initializing the geometric head from pretrained weights and allowing it to adapt during training gives the strongest generation quality, with the best selected video, depth, and point-cloud consistency metrics.
In contrast, fully random initialization causes a clear drop in both video and geometry quality, suggesting that depth supervision alone is not sufficient without a strong geometric prior.
The fixed pretrained geometric head lies between these two settings in generation quality: it retains useful geometric priors and improves substantially over random initialization, but lacks the adaptation capacity of the trainable pretrained geometric head.
We therefore use the trainable pretrained geometric head as the final setting.

\subsection{Qualitative 4D Rollout Visualization}
\label{app:qualitative_4d_rollout}

\begin{figure}[!htbp]
    \centering
    \includegraphics[width=1.0\linewidth]{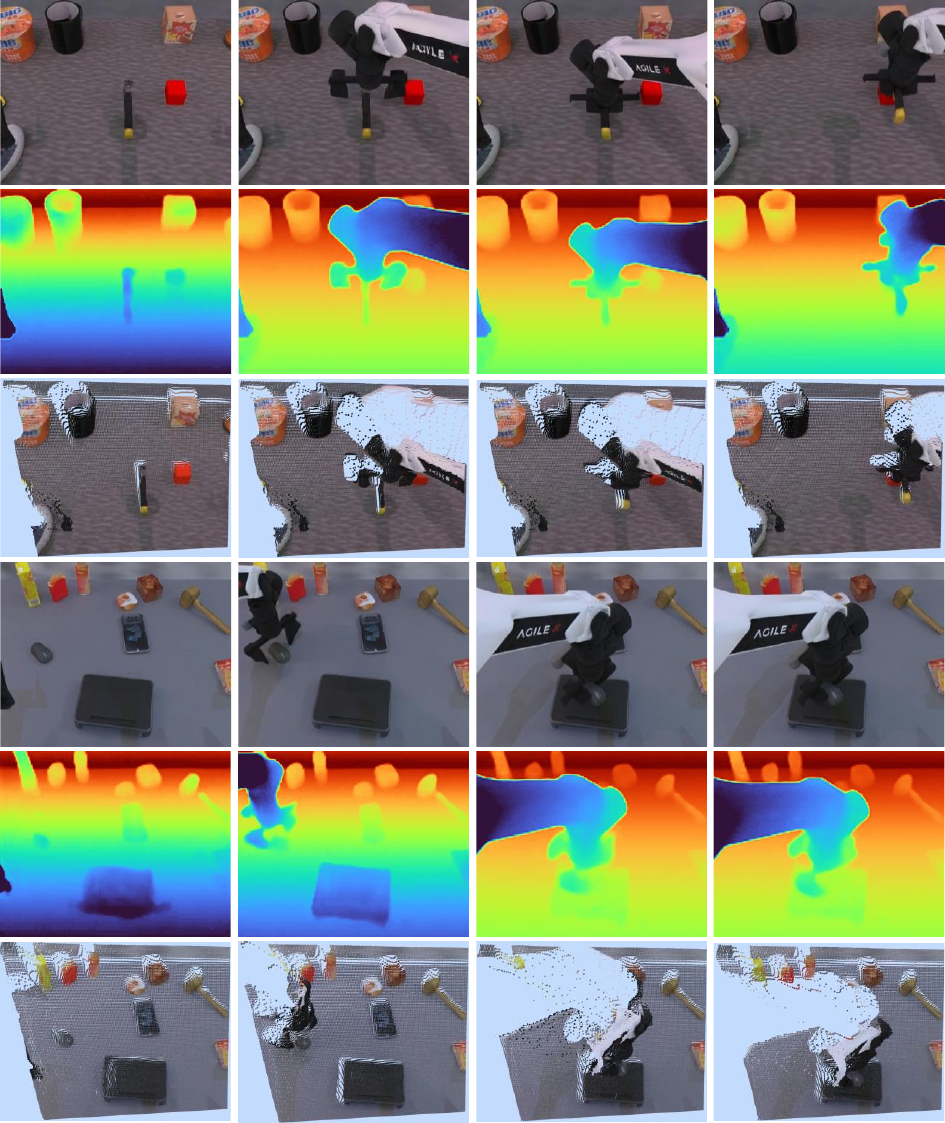}
    \caption{RGB-D and point-cloud rollout visualization. Starting from a single initial frame, \method{} autoregressively predicts future RGB frames and depth maps; the predicted RGB-D frames are then back-projected into point clouds to visualize the induced 4D scene evolution.}
    \label{fig:app_rgbd_rollout}
\end{figure}

Although the spatial-register depth branch is introduced primarily as an auxiliary training objective to regularize causal video features with geometric supervision, it can also be retained for qualitative analysis.
In this setting, \method{} has an interpretable 4D rollout capability: starting from only the first observed frame, the model autoregressively predicts future RGB frames and depth maps, and the generated RGB-D frames can be back-projected into point clouds.
Fig.~\ref{fig:app_rgbd_rollout} visualizes this process.
This analysis path is separate from the default deployment path, where the depth branch is removed and the policy interacts with the environment through lightweight action generation.

\subsection{Failure Analysis}
\label{app:failure_analysis}

\begin{figure}[!htbp]
    \centering
    \includegraphics[width=1.0\linewidth]{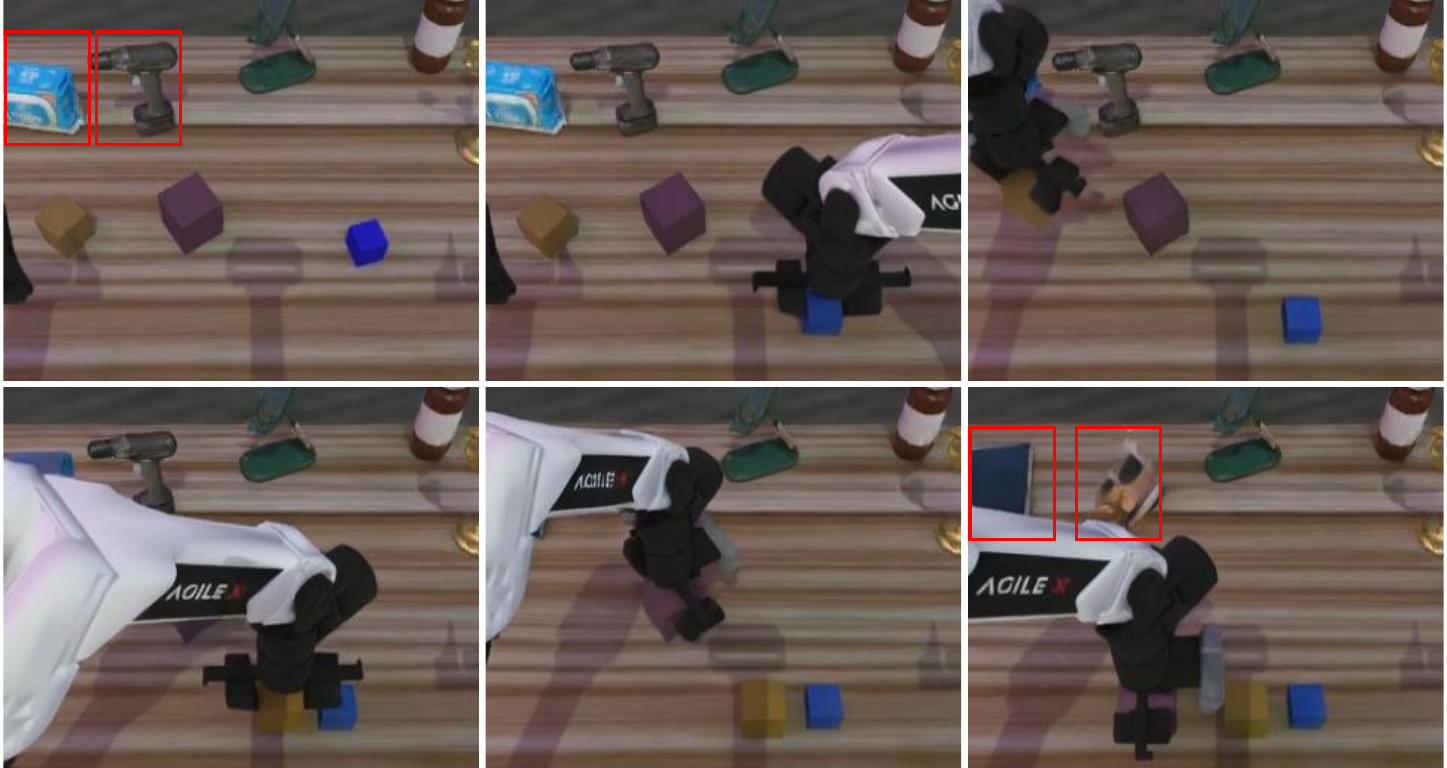}
    \caption{Failure case of long autoregressive rollout. Without an explicit long-term memory, the model may complete an object as a visually plausible but different object after it becomes occluded during rollout.}
    \label{fig:app_failure_case}
\end{figure}

Fig.~\ref{fig:app_failure_case} shows a representative failure case in long autoregressive rollout.
Because \method{} does not introduce an explicit long-term memory, objects that become occluded or leave the visible context may be completed as different objects when the model continues rolling out the scene.
This limitation affects closed-loop visualization of generated futures, but it does not compromise the policy success rate in our evaluation: during control, the model continuously receives fresh observations from the real environment or simulator, rather than relying solely on its own generated rollout.

\FloatBarrier
\subsection{Compute Details}
\label{app:compute}

Tab.~\ref{tab:app_compute} reports the full compute comparison corresponding to the latency and VRAM numbers in Tab.~\ref{tab:robotwin}.
All latency and peak-memory measurements are collected on a single A800 80GB GPU.
Training includes the register blocks and pretrained geometric head. Default \method{} inference removes register tokens, register cross-attention blocks, and the geometric head entirely.

\begin{table}[!h]
    \centering
    \caption{Compute and latency comparison on a single A800 80GB GPU. Latency is reported as mean $\pm$ std in ms when available. Peak memory follows the VRAM measurement in Tab.~\ref{tab:robotwin}.}
    \label{tab:app_compute}
    \scriptsize
    \resizebox{\linewidth}{!}{
    \begin{tabular}{lccc}
        \toprule
        Method / Mode & Denoising steps & Latency / Chunk (ms) & Peak Mem. \\
        \midrule
        $\pi_0$ inference & 10 action steps & $64.16 \pm 0.06$ & 8.45 GiB \\
        $\pi_{0.5}$ inference & 10 action steps & $72.03 \pm 0.06$ & 8.45 GiB \\
        Motus inference & 5 video steps \& 10 action steps & $1516.30 \pm 10.64$ & 11.55 GiB \\
        LingBot-VA inference & 5 video steps \& 10 action steps & $843.57 \pm 11.55$ & 12.97 GiB \\
        Fast-WAM inference & 10 action steps & $425.53 \pm 6.01$ & 11.55 GiB \\
        \method{} inference & 10 action steps & $525.43 \pm 5.64$ & 9.71 GiB \\
        \bottomrule
    \end{tabular}
    }
\end{table}

\FloatBarrier
\section{Conclusion and Limitation}
We present \method{}, a fast 4D world action model that transfers geometric foundation priors into causal video-action representations through spatial register distillation.
Experiments show that \method{} improves spatial consistency and action prediction across simulation and real-world long-horizon tasks.
While WAM is still slower than VLA at present, we aim to further boost its speed in future work.

\method{} improves geometry-aware video-action representations without requiring dense geometry during policy deployment.
However, as discussed in Sec.~\ref{app:failure_analysis}, the model does not maintain an explicit long-term object memory during autoregressive rollout.
This can lead to identity-inconsistent completions after severe occlusion in qualitative generated futures.
Future work could add persistent object memory or scene-state tracking while preserving the lightweight observation-to-action deployment path.

\clearpage
{\small
    \bibliographystyle{corlabbrvnat}
    \bibliography{paper}
}

\end{document}